# Lightweight Residual Network for The Classification of Thyroid Nodules

Nikhila Ponugoti, Sabari Nathan,  Elmer Jeto Gomes Ataide , Alfredo Illanes, Dr. Michael Friebe, Srichandana Abbineni

*Abstract*—Ultrasound is a useful technique for diagnosing thyroid nodules. Benign and malignant nodules that automatically discriminate in the ultrasound pictures can provide diagnostic recommendations or, improve diagnostic accuracy in the absence of specialists. The main issue here is how to collect suitable features for this particular task. We suggest here a technique for extracting features from ultrasound pictures based on the Residual U-net. We attempt to introduce significant semantic characteristics to the classification. Our model gained 95% classification accuracy.

## I. INTRODUCTION

The thyroid is one of the endocrine glands, which generates hormones that assist the body to regulate the metabolism and is situated in the throat just below the epiglottis. Ultrasound scans are most often used for the initial diagnoses of thyroid gland abnormalities. Computer-aided diagnosis (CAD) helps radiologists and physicians improve this diagnostic accuracy, decrease biopsy percentage. Fewer researchers are more focused on Image processing and ROI detections in CAD systems, classify the thyroid nodule is still very difficult because of the feature extraction process. Deep network models are good at image classification and object detection tasks. Here we use a Residual U-Net Encoder part for the classification of ultrasound thyroid nodules. We trained the Residual U-net with randomly initialized weights and learned the encoder features based on a segmentation mask.

## II. METHODS

In this paper, we used 99 images in those 17 benign and 82 malignant images. We augmented the images using different techniques like flip, blur so totally training (1010 images) and testing (590 images). The proposed network has the properties of the encoder and decoder structure of vanilla U-Net [1]. The input image is passed to coordinate the convolution layer [2], the output of this image passed to the encoders. During down-sampling four blocks are used in the encoder phase. In each block, the first layer is a 3×3 convolutional layers, followed by two residual blocks are added and at the end, 2 × 2 max-pooling layers are attached. In the decoder phase, the same blocks have been used except the max-pooling layer which is replaced with an up-sampling layer. Convolutional Block Attention Module (CBAM) attention layer is connected between Encoder and decoder, the output of the decoder is connected with the softmax layer. Connected the attention layer with Global average pooling for the classification. We trained the models into two parts. Part one is trained U-net with Segmentation and Part two trained the encoder as a classifier with previous weight.

## III. RESULTS & DISCUSSION

TABLE I. TABLE RESULT OF TRAINING AND TESTING DATA

|  | Dataset | Accuracy | Sensitivity | Specificity | Dice |
|---|---|---|---|---|---|
| Training | 1010 | 0.945 | 0.875 | 1 | 0.556 |
| Testing | 590 | 0.9507 | 0.883 | 0.997 | 0.5811 |

Light weight model gained the accuracy of 95% on unseen data with 96% F1 score for the classification of benign and malignant nodules. In this model binary cross entropy loss used to update the encoder weights. Segmentation model weights help to obtain the features from the thyroid region and trained model has 1.3million parameters.

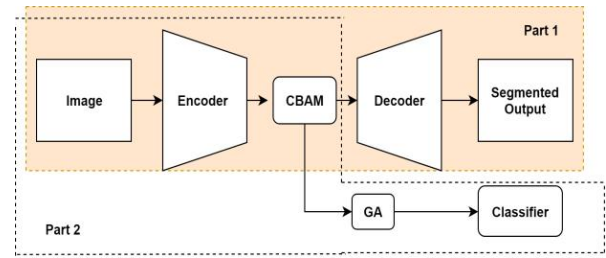

Figure 1. Flow chart of proposed network

## IV. CONCLUSION

Compared to other studies we used the lightweight model to improve the feature extraction process for the thyroid nodule classification and Qing Guan et.al [1] proposed inception-v3 method it exhibits the 90% accuracy on validation data. Compared with Inception-v2 [1] model our model is giving higher accuracy.